\pdfoutput=1
\documentclass[11pt]{article}

\usepackage{acl}

\usepackage{times}
\usepackage{latexsym}

\usepackage[T1]{fontenc}
\usepackage[utf8]{inputenc}

\usepackage{microtype}

\usepackage{inconsolata}

\usepackage{multirow}

\usepackage{amsmath} 
\usepackage{amssymb}
\usepackage{booktabs}

\usepackage{cleveref}
\crefformat{section}{\S#2#1#3}
\crefformat{subsection}{\S#2#1#3}
\crefformat{subsubsection}{\S#2#1#3}
\crefrangeformat{section}{\S\S#3#1#4 to~#5#2#6}
\crefmultiformat{section}{\S\S#2#1#3}{ and~#2#1#3}{, #2#1#3}{ and~#2#1#3}

\title{Related Work and Citation Text Generation: A Survey}

\author{Xiangci Li ~~ Jessica Ouyang\\
  Department of Computer Science \\
  University of Texas at Dallas \\
  Richardson, TX 75080 \\
  \tt lixiangci8@gmail.com, \\ 
  \tt Jessica.Ouyang@UTDallas.edu
}

\begin{document}
\maketitle
\begin{abstract}
To convince readers of the novelty of their research paper, authors must perform a literature review and compose a coherent story that connects and relates prior works to the current work. This challenging nature of literature review writing makes automatic related work generation (RWG) academically and computationally interesting, and also makes it an excellent test bed for examining the capability of SOTA natural language processing (NLP) models. Since the initial proposal of the RWG task, its popularity has waxed and waned, following the capabilities of mainstream NLP approaches. In this work, we %aim to help readers with sufficient basic knowledge about NLP but limited experience with RWG quickly grasp the background, %identify and tackle the limitations of the previous studies, and pinpoint the potential novelty of their studies.
%by surveying 
survey the zoo of RWG historical works, summarizing the key approaches and task definitions and discussing the ongoing challenges of RWG.
\end{abstract}

\section{Introduction}
Academic research is an exploratory activity to solve problems that have never been resolved before. Each academic research paper must sit at the frontier of the field and present novelties that have not been addressed by prior work; to convince readers of the novelty of the current work, the authors must perform a literature review to compare their work with the prior work. In natural language processing (NLP), a short literature review is usually conducted under the “Related Work” section (RWS). Writing an RWS is non-trivial; it is insufficient to simply concatenate generic summaries of prior works. Instead, composing a coherent story that connects each related work and the current (citing) work, reflecting the author's understanding of their field, is preferred \cite{li2024explaining}.

The challenging nature of RWS writing makes automatic related work generation (RWG) an academically and computationally interesting problem. RWG is a complex task that involves multiple NLP subtasks, such as retrieval-augmented generation, long document understanding, and query-focused multi-document summarization.
Moreover, since most NLP papers have an RWS and NLP researchers are natural domain experts for evaluating these RWS, the RWG task is an excellent test bed for examining the capability of SOTA NLP models.

RWG also fills a practical need. Due to the rapid pace of research publications, including pre-prints that have not yet been peer-reviewed, keeping up to date with the latest work in a research area is very time-consuming. Even with daily feed tools, like the Semantic Scholar Research Feed\footnote{https://www.semanticscholar.org/faq/what-are-research-feeds}, researchers still have to curate, read, and digest all the new papers in their feed. Thus, there is a need for concise, automatically generated literature reviews that regularly summarize the papers in a user's feed.

Since \citet{hoang-kan-2010-towards} initially proposed the task, the popularity of RWG has waxed and waned, following the capabilities of mainstream NLP approaches: from rule-based to extractive summarization, then to abstractive summarization on the sentence level, and finally to abstractive section-level RWG. Currently there is a surge of renewed interest in RWG due to the recent success of large language models (LLMs). In this work, we survey the zoo of RWG historical works. 

\begin{table*}[t]
    \scriptsize
     \centering
     \begin{tabular}{l ccc c cc c cc c cc} 
         \toprule
          & \multicolumn{3}{c}{\textbf{Output Unit}} && \multicolumn{2}{c}{\textbf{Cited Paper Input}} && \multicolumn{2}{c}{\textbf{Citation Order/Grouping}} && \multicolumn{2}{c}{\textbf{Availability}} \\ 
          \cmidrule{2-4} \cmidrule{6-7} \cmidrule{9-10} \cmidrule{12-13}
          & \textbf{Sent.} & \textbf{Para.} & \textbf{Sect.} && \textbf{Excerpts} & \textbf{Full Text} && \textbf{Given} & \textbf{Predicted} && \textbf{Code} & \textbf{Data} \\
          \midrule
          \textbf{Extractive} \\
            \cmidrule{1-1}
          \citet{hoang-kan-2010-towards} &&&$\checkmark$ & &&$\checkmark$ & &$\checkmark$& & &&$\checkmark$ \\
          \citet{hu2014automatic} &&&$\checkmark$ & &$\checkmark$& & &&$\checkmark$ & && \\
          \citet{wang-etal-2018-neural-related} &&&$\checkmark$ & &&$\checkmark$ & &$\checkmark$& & &&$\checkmark$ \\
          \citet{chen2019automatic} &&&$\checkmark$ & &$\checkmark$& & &$\checkmark$& & &&  \\
          \citet{wang2019toc} &&&$\checkmark$ & &&$\checkmark$ & &$\checkmark$& & &&$\checkmark$  \\
          \citet{deng2021automatic} &&&$\checkmark$ & &$\checkmark$& & &&$\checkmark$ & && \\
          \midrule
          \textbf{Abstactive (citation)} \\
          \cmidrule{1-1}
          \citet{abura2020automatic} &$\checkmark$&& & &$\checkmark$& & && & &$\checkmark$&$\checkmark$ \\
          \citet{xing2020automatic} &$\checkmark$&& & &$\checkmark$& & && & &&$\checkmark$ \\
          \citet{ge-etal-2021-baco} &$\checkmark$&& & &$\checkmark$& & && & && \\
          \citet{luu-etal-2021-explaining} & $\checkmark$&& & &$\checkmark$& & && & &&$\checkmark$ \\
          \citet{jung2022intent} &&$\checkmark^{*}$& & &$\checkmark$& & && & &$\checkmark$&$\checkmark$ \\
          \citet{li-etal-2022-corwa} &&$\checkmark^{*}$& & &$\checkmark$& & && & & $\checkmark$&$\checkmark$ \\
          \citet{gu2023controllable} &$\checkmark$&& & &$\checkmark$& & && & &$\checkmark$&$\checkmark$ \\
          \citet{li2023cited} &&$\checkmark^{*}$& & &&$\checkmark^{\dag}$ & && & & $\checkmark$&$\checkmark$ \\
          \citet{mandal2024contextualizing} &&$\checkmark^{*}$& & &$\checkmark$& & && & & $\checkmark$&$\checkmark$ \\
            \midrule
          \textbf{Abstractive (section)} \\
          \cmidrule{1-1}
          \citet{chen-etal-2021-capturing} & &$\checkmark$& & &$\checkmark$& & &$\checkmark$& & &&$\checkmark$  \\
          \citet{chen2022target} & &$\checkmark$& & &$\checkmark$& & &$\checkmark$& & &&$\checkmark$  \\
          \citet{liu-etal-2023-causal} &&$\checkmark$& & &$\checkmark$& & &&$\checkmark$ & &&$\checkmark$ \\
          \citet{li2024explaining} &&&$\checkmark$ & &&$\checkmark^{\dag}$ & &$\checkmark$& & &$\checkmark^{\ddag}$ & \\
          \citet{martin2024shallow} &&&$\checkmark$ & &$\checkmark$& & &&$\checkmark^{**}$ & &$\checkmark^{\ddag}$ &$\checkmark$ \\
          \bottomrule
     \end{tabular}
     \caption{Comparison of the task definitions of extractive and both single-citation and full-section abstractive approaches to related work generation. * indicates works that allow multi-sentence citations. $\dag$ indicates works that extract snippets/features from the cited paper full text. ** indicates works that use human editing to improve predicted citation groupings. $\ddag$ indicates works that provide large language model prompts.}
     \label{tab:citation-gen}
 \end{table*}

 We find that, surprisingly, most RWG works are not directly comparable because they vary drastically in task definition and simplifying assumptions (Section \ref{sec:problem_formulation}), as well as using different input features and representations (Section \ref{sec:findings}). There is no standard benchmark dataset for RWG (Section \ref{sec:datasets}), as most works apply custom preprocessing to extract RWS or individual citations, reflecting differences in their task definitions. Further, many works do not release their models or generated outputs, so it is often impossible for later works to compare against earlier approaches (Section \ref{sec:evaluation}). %Nevertheless, we are able to identify common features and auxiliary tasks that benefit the performance of RWG models (Section \ref{sec:findings}). 
Finally, we discuss ethical concerns related to RWG, such as plagiarism and non-factual statements, and the potential consequences of fully automatic RWG on the human process of scientific thinking and writing (Section \ref{sec:ethics}).

%and point out issues and insights of RWG: As the RWG performance drastically increases due to the dramatic success of large language models (LLMs), a practical ethical concern emerges:  users may abuse the RWG system to generate RWS automatically for paper publications, which is a potentially serious violation of academic integrity. Fearing this abusive use, some reviewers reject RWG-related papers and grant submissions because of the legitimacy of RWG research, which didn't happen when RWG was initially proposed. Moreover, there is not yet a standard task definition and benchmark dataset widely acknowledged across different works till today. We further summarize the findings and discuss the future direction of RWG.

\section{Task Definition} \label{sec:problem_formulation}
The task definition for RWG has varied as the SOTA text summarization approach has evolved over time. Even where the overall approach is similar (e.g. extractive or abstractive approaches), different assumptions are made with respect to the availability of system inputs and the unit at which an RWS is generated (Table \ref{tab:citation-gen}).

\subsection{Extractive Related Work Generation}
\label{sec:extractive_formualtion}

\citet{hoang-kan-2010-towards} defined RWG as generating the RWS of a target paper given the rest of the target paper and all cited papers. This focus on extracting and concatenating salient sentences from the cited papers to form an RWS was used by most subsequent extractive RWG approaches \cite{hu2014automatic, wang-etal-2018-neural-related, deng2021automatic}. One key variant is that of \citet{chen2019automatic, wang2019toc}, who extracted sentences from other works that also referenced the cited papers. Otherwise, the main difference among extractive approaches is in how they order the extracted sentences: \citet{hoang-kan-2010-towards, wang-etal-2018-neural-related, chen2019automatic, wang2019toc} assumed the correct ordering as input (either via a human-constructed topic tree or the ground truth ordering of the target RWS), while \citet{hu2014automatic, deng2021automatic} used topic modeling and a sentence reordering module, respectively, to predict an ordering. 

\subsection{Abstractive Related Work Generation}

With the advent of neural language models, two different versions of the abstractive RWG task have been proposed: generating single citation texts versus paragraphs or full RWS. %Detailed inputs and outputs for each prior work are summarized in Appendix Table \ref{tab:task_formulation_abstractive}.

\subsubsection{Citation Text Generation}
Early neural language models, such as the Pointer-Generator \cite{see-etal-2017-get} and early pretrained Transformers \cite{vaswani2017attention}), were capable of fluent abstractive summarization but had severe input length restrictions. Because scientific research papers are very long documents, a new version of the RWG task arose: generating individual citation texts. The system input now needed to include only one or a few cited papers, and to further shorten the system input, researchers no longer included the full texts of the target and cited papers, but used only the target citation context and the cited paper abstract (and occasionally the introduction and conclusion sections).

The main difference among single citation text generation works is in how a citation is defined. \citet{abura2020automatic, xing2020automatic, ge-etal-2021-baco, luu-etal-2021-explaining, gu2023controllable} restrict citation texts to be single sentences; \citet{jung2022intent} allow any number of consecutive sentences, while \citet{li-etal-2022-corwa, li2023cited, mandal2024contextualizing} additionally allow citations that are shorter than a full sentence. Almost all works restrict citations to contain only one cited paper; only \citet{li-etal-2022-corwa, li2023cited, mandal2024contextualizing} explicitly allow multiple cited papers.

%given the rest of the available contexts: the context RWS texts before or after the target citation texts, the full text of the target paper except the target RWS, the full list of full texts of the cited papers etc. This target shift happened because empirically the fine-tuned neural network models were not able to reliably generate section-level RWS.

\subsubsection{Section-Level Generation}

\citet{chen-etal-2021-capturing, chen2022target} pioneered section-level RWG by treating the paragraph as the unit of generation; they required that a target paragraph contain at least two citations, explicitly distinguishing their work from the single citation text generation setting. While \citet{chen-etal-2021-capturing, chen2022target} used the given paragraph organization of the target RWS, subsequent works focused on ordering and organizing citations into paragraphs and generating transitional sentences between citations \cite{liu-etal-2023-causal,li2024explaining,martin2024shallow}. 

Further, the great success of SOTA LLMs in multiple natural language understanding and generation tasks, combined with their large context windows, have recently made it possible to generate a full RWS in a single pass \cite{li2024explaining,martin2024shallow}. Thus, the task definition has now returned to the full RWS generation originally proposed by \citet{hoang-kan-2010-towards} and previously tackled only by extractive approaches.

\setlength{\tabcolsep}{5pt}

\begin{table*}[t]
    \scriptsize
     \centering
     \begin{tabular}{l ccccc c cccc c ccc} 
         \toprule
          & \multicolumn{5}{c}{\textbf{Cited Paper Representation}*} && \multicolumn{4}{c}{\textbf{Target Paper Context}} && \multicolumn{3}{c}{\textbf{Citation Analysis Use}} \\ 
          \cmidrule{2-6} \cmidrule{8-11} \cmidrule{13-15}
          & \textbf{Intro.} & \textbf{RWS} & \textbf{Concl.} & \textbf{CTS} & \textbf{Graph} && \textbf{Abs.} &\textbf{Intro.} & \textbf{RWS$^{\dag}$} & \textbf{Concl.} && \textbf{MTL} & \textbf{Control} & \textbf{Eval.} \\
          \midrule
          \textbf{Extractive} \\
            \cmidrule{1-1}
          \citet{hoang-kan-2010-towards} &$\checkmark$&$\checkmark$&$\checkmark$&& & &&&& & &&& \\
          \citet{hu2014automatic} &$\checkmark$&$\checkmark$&$\checkmark$&& & &$\checkmark$&$\checkmark$&& & &&& \\
          \citet{wang-etal-2018-neural-related} &$\checkmark$&$\checkmark$&$\checkmark$&&$\checkmark$ & &&&& & &&& \\
          \citet{chen2019automatic} &$\checkmark$&&$\checkmark$&& & &$\checkmark$&$\checkmark$&&$\checkmark$ & &&& \\
          \citet{wang2019toc} &$\checkmark$&$\checkmark$&$\checkmark$&$\checkmark$& & &$\checkmark$&$\checkmark$&$\checkmark$&$\checkmark$ & &&& \\
          \citet{deng2021automatic} &&&$\checkmark$&& & &&&& & &&& \\
          \midrule
          \textbf{Abstactive (citation)} \\
          \cmidrule{1-1}
          \citet{abura2020automatic} &&&&& & &&&& & &&& \\
          \citet{xing2020automatic} &&&&& & &&&$\checkmark$& & &&& \\
          \citet{ge-etal-2021-baco} &&&&&$\checkmark$ & &&&$\checkmark$& & &$\checkmark$&& \\
          \citet{luu-etal-2021-explaining} &&&&& & &&$\checkmark$&& & &&& \\
          \citet{jung2022intent} &&&&& & &$\checkmark$&&& & &&$\checkmark$& \\
          \citet{li-etal-2022-corwa} &&&&& & &&&$\checkmark$& & &&& \\
          \citet{gu2023controllable} &&&&& & &$\checkmark$&&& & &&$\checkmark$& \\
          \citet{li2023cited} &&&&$\checkmark$& & &&&$\checkmark$& & &&&\\
          \citet{mandal2024contextualizing} &&&&& & &&&$\checkmark$& & &&& \\
            \midrule
          \textbf{Abstractive (section)} \\
          \cmidrule{1-1}
          \citet{chen-etal-2021-capturing} &&&&&$\checkmark$ & &&&& & &&& \\
          \citet{chen2022target} &&&&&$\checkmark$ & &$\checkmark$&&& & &&& \\          
          \citet{liu-etal-2023-causal} &&&&&$\checkmark$ & &&&& & &&& \\
          \citet{li2024explaining} &$\checkmark^{**}$&$\checkmark^{**}$&$\checkmark^{**}$&$\checkmark$&$\checkmark$ & &$\checkmark$&$\checkmark$&$\checkmark$& & &&$\checkmark$&$\checkmark$ \\
          \citet{martin2024shallow} &&&&& & &$\checkmark$&$\checkmark$&$\checkmark$& & &&&$\checkmark$ \\
          \bottomrule
     \end{tabular}
     \caption{Comparison of RWG approaches. * All surveyed works used cited paper titles and abstracts, which are not list in this table. $\dag$ The target citation itself is masked. $\checkmark^{**}$ indicates features extracted from the listed sections.}
     \label{tab:overview}
 \end{table*}

\section{Overview of Approaches}
\label{sec:findings}
%\subsection{A Brief History}
%The history of RWG can be roughly divided into three eras: extractive RWG, abstractive sentence-level RWG, and abstractive section-level RWG. The approaches by each prior work are summarized in Appendix Table \ref{tab:approaches}.
When \citet{hoang-kan-2010-towards} proposed the RWG task, they identified three main steps: (1) Finding relevant documents, (2) Identifying the salient aspects of these documents with respect to the current work; (3) Generating a topic-biased summary. In practice, all existing works skip the document retrieval step by using the gold cited paper list in the target RWS. At a high level, the methodologies of most extractive, citation-level and section-level abstractive RWG approaches are similar within their respective categories: extractive approaches focus on the salience step and simply concatenate the extracted sentences to form the summary, while abstractive approaches focus on directly generating the summary, often without explicitly modeling salience. In this section, we do not give an exhaustive description of all methodologies, but highlight some common features and design perspectives from the overall body of RWG work (summarized in Table \ref{tab:overview}). The details of individual works can be found in Appendix \ref{sec:appendix}.

%Throughout the retrospective studies of RWG prior works, we generally observe that RWG modeling approaches closely follow and evolve as the SOTA models since every new generation of NLP models (e.g. rule-based, graphical model, custom neural network, BERT-family, LLMs etc.) beats all variations of the previous generation of models. While the true improvement originates from non-model factors such as the improved usage of input features. 
%In other words, if the initial RWG work \cite{hoang2010towards} had access to today's SOTA models, e.g. LLMs, the only possible difference may come from the usage of features. 
%In this section, we summarize common key factors that improve RWG system performance.

\subsection{Representing Cited Papers}

\textbf{Abstracts.} In abstractive RWG approaches, and some extractive approaches, the cited paper title and abstract are commonly used as a proxy for its full text \cite{abura2020automatic, xing2020automatic, ge-etal-2021-baco, chen-etal-2021-capturing, chen2022target, jung2022intent, li-etal-2022-corwa, gu2023controllable, liu-etal-2023-causal, mandal2024contextualizing, martin2024shallow}, occasionally augmented with the introduction and/or conclusion \cite{hu2014automatic, chen2019automatic, deng2021automatic}. 

The abstract is a concise summary of the central ideas of the cited paper and can fit in a neural language model's input length limit where the full text cannot. Abstracts also play an important role in scientific communication as a preview of the paper, so they are easy to access even when their fulls text are blocked by paywalls. \citet{li2024explaining} find that generated RWS conditioned on cited paper abstracts are preferred by human readers over those conditioned on LLM-generated faceted summaries \cite{meng-etal-2021-bringing} of the cited papers.

\textbf{Cited Text Spans (CTS).} \citet{li2023cited} proposed to condition on automatically predicted CTS rather than cited paper abstracts. CTS refers to the specific span of the cited paper that a given citation refers to; to draw a parallel to claim verification, the citation can be thought of as the claim, and the CTS as its supporting evidence. Thus, \citet{li2023cited} effectively proposed an extract-then-abstract approach to citation text generation, arguing that the cited paper abstract may not always contain sufficient information to ground the target citation. It is interesting to note that CTS had previously been used for extractive RWG by \citet{wang2019toc}, who extracted CTS for other citations of the cited paper in works similar to the target paper.

\textbf{Citation Graphs.} Since an RWS describes the relationship between the target paper and prior work, as well as among prior works, some section-level RWG approaches have modeled the local citation network of the target and cited papers. \citet{wang-etal-2018-neural-related} used a random walk on a heterogeneous bibliography graph consisting of paper, author, venue, and keyword nodes to prune the search space of salient sentences for extractive RWG. \citet{ge-etal-2021-baco, chen-etal-2021-capturing, chen2022target} used customized neural network architectures inspired by Graph Attention Networks \cite{Velickovic2018GraphAN} to encode the local citation network as an additional input for abstractive RWG, while \citet{li2024explaining} prompted an LLM to generate a natural language description of the relationship between a pair of papers in the citation network.

\subsection{The Importance of Citation Context}
Citation context refers to the text preceding or surrounding the target citation or RWS. In the case of individual citations, the context is commonly defined as several sentences before, and optionally after, the target citation \cite{xing2020automatic, ge-etal-2021-baco, li-etal-2022-corwa, li2023cited, mandal2024contextualizing}; for some citation text generation works and most section-level RWG works, the context can be the full text of the target paper, or a few key sections, most commonly the title, abstract, introduction, and conclusion \cite{luu-etal-2021-explaining, jung2022intent, gu2023controllable, chen2022target, li2024explaining, martin2024shallow}.

Intuitively, the context indicates which topics are salient to the target paper, restricting the RWG solution space. Extractive works \cite{hu2014automatic, chen2019automatic, wang2019toc} used the context as a query to score cited paper sentences.
%Since extractive RWG approaches barely modify the output RWS, citation contexts are more important for abstraction approaches. 
In abstractive approaches, conditioning on the context improves the coherence of the generated text with the rest of the target paper; \citet{mandal2024contextualizing} found human readers preferred citations generated using the entire context, with the target citation embedded inside it, as the generation target.

It is interesting to note that a few works did not use any target paper context at all \cite{hoang-kan-2010-towards, abura2020automatic, chen-etal-2021-capturing}, but these were early works in their respective categories (extractive versus abstractive citation- or section-level generation), and later works all used target paper context.

\subsection{Applying Citation Analysis}

Citation analysis is a related area of research studying the properties of citations in scientific writing. Several studies have proposed taxonomies such as citation function \cite{garfield1965can, teufel2006automatic, dong2011ensemble, jurgens2018measuring, tuarob2019automatic, zhao2019context}, citation intent \cite{cohan2019structural, lauscher2021multicite}, and citation sentiment \cite{athar2011sentiment, athar-teufel-2012-context, ravi2018article}, and such labels have been used to improve RWG performance. 

\citet{ge-etal-2021-baco} used citation function prediction as an auxiliary training objective. \citet{jung2022intent, gu2023controllable} used citation intents to perform controllable citation text generation. Inspired by the observation of \citet{lauscher-etal-2022-multicite} that simple citation label sets struggle to represent ambiguous, real-world citations, \citet{li2024explaining} used LLM-generated, natural language descriptions of function of a cited paper in other, similar works that also cited it.

%As Section \ref{sec:main_story} elaborates, RWG systems should be ultimately optimized to fit the end-user preferences, not the evaluation metrics. Similar to the backbone story, writing style is another key factor that humans are sensitive to. Because previous extractive RWG did not make major edits to the extracted sentences, and citation text generation works mainly aimed to improve the textual overlap of the predicted texts and the gold texts, the writing style was neglected for a long time. As LLMs become capable of generating human-like passages, writing style emerges as one of the important factors to evaluate.

Other work has studied the discourse properties and organization of citations. \citet{jaidka2010imitating, jaidka2011literature, khoo2011analysis, jaidka2013literature, jaidka2013deconstructing} classified literature reviews into \textit{integrative} (summarizing individual cited papers) and \textit{descriptive} (focusing on high-level ideas from multiple papers) writing styles. \citet{li-etal-2022-corwa} proposed a more fine-grained taxonomy at the citation level, labeling citations as \textit{dominant} (the main focus of their sentence) or \textit{reference} (tangential to the rest of their sentence).%, and trained separate generation models to target each type. XL: This is not true!
%Intuitively, integrative literature reviews have significantly more research result information and critiques than descriptive literature reviews, while descriptive literature reviews have more research method information introduced. Interestingly, by studying the sections that each type of literature reviews cite, they find that integrative literature reviews reference more information from the results and conclusion sections than descriptive literature reviews. On the other hand, descriptive literature reviews reference more information from the abstract and introduction sections than integrative literature reviews. By comparing the type of information transformation from the cited paper to the literature review sentences used in these two types of literature reviews, including cut-paste, paraphrase, summary, and critical reference, they find that integrative literature reviews have more critique, high-level summarizing, and paraphrasing of source information than descriptive literature reviews. On contrary, descriptive literature reviews have more cut-pasting than integrative literature reviews. This characteristic also holds for the abstract sections that both styles of literature reviews frequently reference. Integrative literature reviews are likely to paraphrase information from the abstract, while descriptive literature reviews are likely to cut-paste information from the abstract. Finally, they find that a large proportion of the research objectives and research methods sentences in either style of the literature review are summarized at a high level.

\citet{li2024explaining} used this taxonomy to analyze the writing style of LLM-generated RWS and observed a strong correlation between the proportion of reference-type citations and human preference scores, concluding that human readers prefer integrative RWS supported by reference-type citations. Similarly, \citet{martin2024shallow} found that both human-written and human-assisted, LLM-generated RWS had significantly more cited papers per sentence than pure machine-generated RWS.

\subsection{Human-Assisted Generation}  \label{sec:main_story}
%RWG is a task with a large search space \cite{li-etal-2022-corwa}. In other words, there exist multiple plausible RWS given the same citation texts and reference list. 
While RWG models are optimized to reconstruct the original citation texts or RWS in their training datasets, the ultimate goal of the task is to generate an RWS that satisfies a user. Human readers are sensitive to errors in cited paper organization (e.g. papers cited in the same paragraph are not sufficiently related to each other) and emphasis (e.g. less salient papers are described in greater detail than more salient ones); currently, even SOTA LLMs are not capable of organizing and emphasizing a set of cited papers without human guidance \cite{li2024explaining, martin2024shallow}.

%Since it's unnecessary and impractical to fully automatically predict the exact text that the end user desires, the human-in-the-loop approach is a good workaround. 
Thus, human input has been included in several RWG works. To determine the most salient aspects of a cited paper for single citation text generation, \citet{li2023cited} proposed to retrieve cited text spans (CTS) using user-provided keywords as queries, while %This also addressed the problem that all prior CTS-based works \cite{yasunaga2017graph, wang2019toc} retrieved CTS with gold citations, which does not exist in the real scenario. 
\citet{gu2023controllable} directly used human-written keywords as an additional input. \citet{li2024explaining} extended this idea to section-level RWG by proposing to use a human-written short summary of the main ideas of the target RWS. Also for section-level RWG, \citet{martin2024shallow} introduced a human-in-the-loop component where the user edited an predicted cited paper grouping before the generation step.

\section{Datasets}
\label{sec:datasets}
Despite the twenty published works on RWG, there is no standard benchmark dataset for the task. As we discussed in Section \ref{sec:problem_formulation}, most RWG works define their own version of the task; they also create their own datasets, adapted to their particular task definition. In this section, we describe the most commonly used sources of scientific articles (Table \ref{tab:dataset-summary}) and summarize how RWG works have built on these sources. The details of each work's datasets can be found in Appendix Table \ref{tab:datasets}.

\begin{table*}[t]
    \scriptsize
     \centering
     \begin{tabular}{l cccc c ccc c} 
         \toprule
          & \multicolumn{4}{c}{\textbf{Data Source}} && \multicolumn{3}{c}{\textbf{Domain}} \\ 
          \cmidrule{2-5} \cmidrule{7-9}
          & \textbf{AAN} & \textbf{S2ORC} & \textbf{Delve} & \textbf{Other} && \textbf{NLP/AI Only} & \textbf{General CS} & \textbf{Non-CS} & \textbf{Available?} \\
          \midrule
          \textbf{Extractive} \\
            \cmidrule{1-1}
          \citet{hoang-kan-2010-towards} &&&&$\checkmark$ & &$\checkmark$&& &$\checkmark^{\dag}$ \\
          \citet{hu2014automatic} &$\checkmark$&&& & &$\checkmark$&& \\
          \citet{wang-etal-2018-neural-related} &&&&$\checkmark$ & &&$\checkmark$& &$\checkmark$ \\
          \citet{chen2019automatic} &$\checkmark$&&&$\checkmark$ & &$\checkmark$&& \\
          \citet{wang2019toc} &&&&$\checkmark$ & &$\checkmark$&&  &$\checkmark$ \\
          \citet{deng2021automatic} &$\checkmark^{*}$&&& & &$\checkmark$&& \\
          \midrule
          \textbf{Abstactive (citation)} \\
          \cmidrule{1-1}
          \citet{abura2020automatic} &$\checkmark^{*}$&&& & &$\checkmark$&&  &$\checkmark$ \\
          \citet{xing2020automatic} &$\checkmark$&&& & &$\checkmark$&&  &$\checkmark$ \\
          \citet{ge-etal-2021-baco} &$\checkmark$&&& & &$\checkmark$&& \\
          \citet{luu-etal-2021-explaining} &&$\checkmark$&& & &&$\checkmark$&  &$\checkmark$ \\
          \citet{jung2022intent} &&&&$\checkmark$ & &&$\checkmark$&$\checkmark$ &$\checkmark$ \\ % SciCite (citation intent dataset)
          \citet{li-etal-2022-corwa} &&$\checkmark^{**}$&& & &$\checkmark$&&  &$\checkmark$ \\
          \citet{gu2023controllable} &&&&$\checkmark$ & &&$\checkmark$& &$\checkmark$ \\
          \citet{li2023cited} &&$\checkmark^{**}$&& & &$\checkmark$&& &$\checkmark$ \\
          \citet{mandal2024contextualizing} &&$\checkmark^{**}$&& & &$\checkmark$&& &$\checkmark$ \\
            \midrule
          \textbf{Abstractive (section)} \\
          \cmidrule{1-1}
          \citet{chen-etal-2021-capturing} &&$\checkmark$&$\checkmark$& & &&$\checkmark$&$\checkmark$ &$\checkmark$ \\
          \citet{chen2022target} &&$\checkmark$&$\checkmark$& & &&$\checkmark$&$\checkmark$ &$\checkmark$ \\
          \citet{liu-etal-2023-causal} &&$\checkmark$&$\checkmark$& & &&$\checkmark$&$\checkmark$ &$\checkmark$ \\
          \citet{li2024explaining} &&&&$\checkmark$ & &&$\checkmark$&$\checkmark$ \\
          \citet{martin2024shallow} &&&&$\checkmark$ & &$\checkmark$&& &$\checkmark$ \\
          \bottomrule
     \end{tabular}
     \caption{List of common datasets used in related work generation. * indicates works that use the SciSummNet subset of AAN. ** indicates works that use the CORWA subset of S2ORC. $\dag$ indicates works that published their datasets, but the repositories are no longer accessible.}
     \label{tab:dataset-summary}
 \end{table*}
 
% No data:
% \cite{luu-etal-2021-explaining}

% No model:
% \cite{hoang2010towards, chen2019automatic, xing-etal-2020-automatic, chen-etal-2021-capturing}

% Neither:
% \cite{hu2014automatic, wang2019toc, deng2021automatic, ge-etal-2021-baco}

\subsection{Common Datasets}

\paragraph{The ACL Anthology Network (AAN) Corpus} 
\cite{radev2013acl} consists of papers published by the Association for Computational Linguistics (ACL). %It has been used by many of the surveyed papers.
For each paper, it annotates the set of sentences in any other AAN paper that cite that paper. Both in the construction of AAN, as well as in single citation text generation works that use it, individual citation texts are extracted via string search for citation marks, such as ``Smith et al. (2024)" or ``[1]" \cite{xing2020automatic, ge-etal-2021-baco}.

\paragraph{SciSummNet} \cite{yasunaga2019scisummnet}, used by \citet{abura2020automatic, deng2021automatic}, is a subset 1000 papers from the AAN Corpus with human-validated citation sentences and summaries.

\paragraph{Delve}
\cite{akujuobi2017delve} consists of papers from several computer science conferences spanning multiple fields of research. It includes automatically extracted paper abstracts and full text, as well as citation texts and links.

\paragraph{The Semantic Scholar Open Research Corpus (S2ORC)} 
\cite{lo-wang-2020-s2orc} contains open-access papers from multiple disciplines. The papers are annotated with automatically detected inline mentions of citations, figures, and tables, which saves researchers the need to process raw PDF files. 
%S2ORC is constructed using data from the Semantic Scholar literature corpus \cite{ammar2018construction}, which derives papers from numerous sources: obtained directly from publishers, from resources such as Microsoft Academic Graph (MAG) \cite{shen2018web}, from various archives such as arXiv or PubMed, or crawled from the open Internet in the format of PDFs or LaTeX. Thus S2ORC consists of a large number of parsed paper texts from various fields of study. %, including medicine, biology, physics, mathematics, computer science, chemistry, psychology, engineering, etc. in decreasing order. 
%Since each inline citation, figure and table are annotated, S2ORC greatly reduces the workload of researchers who work on scientific document processing, by saving the hassle of collecting and pre-processing the raw PDF files. %As \citet{chen-etal-2021-capturing, luu-etal-2021-explaining} use S2ORC for their studies. we believe more and more future works will develop their work based on S2ORC.
%\citet{lo-wang-2020-s2orc} also released their paper parsing tool, doc2json, to parse the papers in PDF to JSON format, so that new papers not available in S2ORC can still be easily processed.

\paragraph{Citation Oriented Related Work Annotation (CORWA)} \cite{li-etal-2022-corwa} is derived from the ACL partition of S2ORC and is annotated specifically for citation text generation. CORWA labels citations and their discourse roles (\textit{dominant} or \textit{reference}).

\subsection{Discussion}
One common challenge with all existing datasets is that, for a given target paper, not all of its cited papers are necessarily in the dataset (e.g. because they are behind a paywall). In single citation text generation works, such missing cited papers are simply omitted from training and testing. For section-level RWG, missing cited papers are a bigger problem, as their absence may disrupt the flow of the generated RWS \cite{li2024explaining}.

It is also interesting to note that the majority of RWG works have used NLP datasets, and almost no works use papers from outside the domain of computer science. It is likely that RWG researchers prefer to use NLP papers because they include a separate RWS that is easy to extract, which is not the case in all fields of research; they are within the researchers' own domain of expertise, making system development easier; and they are in the domain of the researchers' colleagues, making it easier to recruit human judges for evaluation.

Finally, with the advent of LLM-based approaches, RWG researchers must contend with the possibility that a target paper was part of the training data of their model. As a result, LLM-based works have explicitly targeted recent papers \cite{li2024explaining, martin2024shallow}.

\section{Evaluation}
\label{sec:evaluation}

\subsection{Baselines}

As Appendix Tables \ref{tab:extractive_evaluation} \& \ref{tab:abstractive_evaluation} show, there are a few baselines widely used across RWG works. Extractive works commonly use LEAD \cite{wasson1998using}, MEAD \cite{radev2004centroid}, LexRank \cite{erkan2004lexrank}, and TexRank \cite{mihalcea2004textrank}, while abstractive works use naive sequence-to-sequence approaches, with base models such as PTGEN \cite{see-etal-2017-get}, BertSumAbs \cite{liu-lapata-2019-text}, and Longformer Encoder-Decoder \cite{Beltagy2020Longformer}. These common baselines are relatively easy to replicate because they are well-documented, general-purpose summarization approaches. In contrast, most specialized RWG approaches are not easy to replicate and are thus rarely used as baselines for later works; we discuss this issue further in Section \ref{sec:fragmented}.

\subsection{Metrics}

Almost all RWG works use the summarization metric ROUGE \cite{lin2004rouge} as their automatic evaluation metric; \citet{luu-etal-2021-explaining} additionally use the translation metric BLEU \cite{papineni2002bleu}. %Although the ROUGE scores are not directly comparable across studies since different studies use different settings, such as the task definition and dataset, it's still clear that extractive approaches yield much higher ROUGE scores than abstractive methods, since the extracted sentences may come from the original cited texts. 

Most works additionally conduct human evaluations, as is common in natural language generation tasks. While there is no fixed standard for how to conduct an RWG human evaluation, most works evaluate at least 15 samples, with three human judges per sample. Judges are generally asked to rate the \textit{fluency} or \textit{readability}, the \textit{coherence} with respect to the target paper, and the \textit{relevance} or \textit{informativeness} with respect to the cited paper on a five-point Likert scale.

The relatively small number of human-evaluated samples in RWG works is likely due to the difficulty of recruiting human judges with the expertise to understand the generated citation texts or RWS, as well as the high time commitment and difficulty of the task, which requires judges to read multiple, highly specialized documents. A more detailed summary of metrics used in RWG works can be found in Appendix Table \ref{tab:human_evaluation}.

%Because evaluating an RWS is intellectually challenging, there is not yet a solid standard of how to conduct the human evaluation. A general strategy is to sample at least 15 examples and get them rated by three evaluators. There are also no fixed perspectives for human evaluation, but generally, each work evaluates the fluency (readability) of the examples, whether the citation text is coherent with the citing paper's context, and the correctness of the content (relevance) with respect to the cited papers. In general, a 5-point scale is the most popular. %, but 3-point scale \cite{chen-etal-2021-capturing} and yes/no judgment \cite{luu-etal-2021-explaining} is also employed. 

\section{Conclusions and Discussion}

Having surveyed the field of RWG from the perspectives of task definition, approach, datasets, and evaluation methods, we conclude by identifying three main challenges in modern RWG and make recommendations for future work in this area.

\subsection{Lack of Comparability} \label{sec:fragmented}
%Other than the ethical concern, 
Work in RWG is fragmented in terms of task definitions, datasets used for training and evaluation, and how evaluations are conducted. Unlike most NLP tasks, there are no standard benchmarks for RWG. Table \ref{tab:citation-gen} shows that around half of existing works do not release their models or generated citation texts/RWS, making it impossible to reproduce or directly compare approaches. 

As we discuss in Section \ref{sec:problem_formulation}, RWG works do not agree on the definition of \textit{citation} (one or more cited papers discussed in one or more sentences, or just part of a sentence) or \textit{related work section} (a concatenation of individual citations or paragraphs versus one continuous and coherent piece of text). Thus, the target outputs of most RWG systems are not directly comparable to those of other systems.

A deeper problem with the varying definitions of \textit{citation} is noted by \citet{li-etal-2022-corwa}, who argue that human annotators can easily find examples of human-written citations that are longer or shorter than a single sentence, or that contain more than one cited paper, so ignoring citations that are longer than a single sentence or discuss more than a single cited paper is unrealistic. They further argue that restricting citations to be single sentences is problematic when the approach uses citation context; in the case of a multi-sentence citation, an RWG system that assumes each citation can only be one sentence and uses the surrounding sentences as context will actually use the rest of the sentences from the target citation as context, creating an information leakage problem.

Variation in datasets comes partly from differences in the task definition and partly from the fact that, of the commonly used source corpora, only the CORWA partition of S2ORC \cite{li-etal-2022-corwa} is explicitly designed for RWG; the others (AAN, S2ORC, and Delve) are general-purpose scholarly document and citation analysis datasets. As a result, these other source corpora either automatically extract citations by searching for sentences containing citation marks or do not label citations at all; in the latter case, RWG researchers extract citations themselves by searching for sentences containing citation marks and imposing assumptions about the number of cited papers a citation can contain. Besides CORWA, only the annotations of \citet{xing2020automatic} provide human-labeled citations.

Finally, variation in evaluation stems from the existing problem in general summarization research where automated metrics, such as the commonly used ROUGE scores, do not correlate well with human judgments, so many RWG works perform human evaluation. While \textit{fluency}, \textit{coherence}, and \textit{relevance} are commonly used aspects of human evaluation (Appendix Table \ref{tab:human_evaluation}), many works define custom aspects, such as succinctness \cite{chen-etal-2021-capturing, deng2021automatic, liu-etal-2023-causal}, factual correctness \cite{li2024explaining}, and correctness of citation intent \cite{jung2022intent, gu2023controllable}.

\subsection{Common Limitations}

We find several limitations common to existing work on RWG for future work to consider.

\paragraph{Citation ordering and organization.} 
Out of twenty surveyed RWG works, only four attempt to predict the correct ordering and/or grouping of citations into paragraphs \cite{hu2014automatic, deng2021automatic, liu-etal-2023-causal, martin2024shallow}; an additional two papers acknowledge the citation ordering and grouping problem but assume a human-provided ordering is available \cite{hoang-kan-2010-towards} or use a chronological ordering heuristic \cite{li2024explaining}. Yet \citet{li2024explaining} observed that human readers noticed and disliked errors in citation grouping, such as when chronologically adjacent cited papers about different topics were placed in the same paragraph, and \cite{martin2024shallow} found significant differences in the organization of generated RWS with and without human-assisted citation grouping. We suggest fully automatic citation ordering and grouping as an important area for further investigation.

\paragraph{Transition sentences and writing style.} 
Based on the terms from general summarization \cite{klavans2001domain}, \citet{hoang-kan-2010-towards} distinguished \textit{informative} sentences, which ``give detail on a specific aspect of the problem\ldots definitions, purpose or application of the topic", and \textit{indicative} sentences, which ``make the topic transition explicit and rhetorically sound". However, modern abstractive approaches have focused on informative sentences: single citation generation approaches completely ignore indicative transition sentences, and section-level approaches include them only in that they are part of the target paragraphs. \citet{li2024explaining} found that human readers asked for more transition sentences, complaining about RWS that simply concatenated one cited paper summary after another. Further, their analysis of RWS writing style and the citation clusters of \citet{martin2024shallow} have shown that generated RWS do not draw enough connections among cited papers. Thus, the generation of transition sentences and multi-paper citations remains an open problem.

\paragraph{Retrieval-augmented related work generation.}
Existing RWG works assume the list of cited papers is available as input, but this assumption is unrealistic, as evidenced by the existence of ``missing citations" questions on many conference and journal peer review forms. \citet{li2024explaining} reported that several human judges expressed the desire for a system that would not only help them draft a RWS, but also alert them to any other relevant papers they should consider citing. Given the recent success and popularity of retrieval-augmented generation (RAG) approaches \cite{lewis2020retrieval, shuster-etal-2021-retrieval-augmentation}, applying RAG to RWG is a promising direction for future RWG research.

\subsection{Ethical Concerns}
\label{sec:ethics}

Finally, we discuss three ethical issues related to the RWG task. First, abstractive RWG works must be concerned with the problems of plagiarism and factual errors. In extractive approaches, the generated RWS is by its very nature plagiarized, since its sentences are copied directly from the cited papers; it was presumably well-understood by extractive RWG researchers that their systems could never be used to directly write the RWS for a new paper. However, extractive approaches cannot hallucinate, so their outputs are less likely to contain factual errors about the cited papers. 

With modern abstractive RWG, the situation is muddier. It is well-known in general summarization research that abstractive models can still copy significant chunks of text directly from their inputs \cite{grusky-etal-2018-newsroom, xsum-emnlp}, and factual consistency in summarization is an active research area \cite{cao2018faithful, 10.1145/3292500.3330955, falke-etal-2019-ranking, kryscinski-etal-2019-neural}. Thus, it is possible for an abstractive RWG system to output plagiarized or hallucinated text, which should be of concern to any user who wishes to use such a system to write an RWS.

Second, the use of RWG to write an RWS for a paper one intends to submit for publication raises questions of academic dishonesty. Is it ethical for a researcher to put an automatically generated RWS in a submitted manuscript? Does this mean the researcher is claiming to have written that RWS, as they presumably wrote the rest of the paper? Do the answers to these questions change if the researcher has edited the automatically generated RWS? As with many concerns relating to the use of powerful modern LLMs, these questions are very new, and there is as yet no consensus among the scientific community on how to answer them. While automatically generated RWS as currently easy to recognize, we nonetheless urge caution on the part of RWG researchers and users.

Third, RWG is a challenging task even for humans; in many doctoral programs, writing a formal literature review is part of their candidacy qualifying exams \cite{knopf2006doing}. Thus, the process of writing an RWS may be considered an important process for researchers where they must read broadly and think deeply about how their contributions fit into the bigger picture of their field. Some RWG works have argued that writing an RWS is arduous and time-consuming, and so RWG should save researchers from having to do it, but we argue this position ignores the value of RWS writing as a learning and thinking experience. We urge RWG researchers to consider human-in-the-loop frameworks, following \cite{gu2023controllable, li2024explaining} and especially \cite{martin2024shallow}.

\clearpage
\bibliography{custom,anthology}

\begin{thebibliography}{63}
\expandafter\ifx\csname natexlab\endcsname\relax\def\natexlab#1{#1}\fi

\bibitem[{AbuRa’ed et~al.(2020)AbuRa’ed, Saggion, Shvets, and Bravo}]{abura2020automatic}
Ahmed AbuRa’ed, Horacio Saggion, Alexander Shvets, and Alex Bravo. 2020.
\newblock Automatic related work section generation: experiments in scientific document abstracting.
\newblock \emph{Scientometrics}, 125(3):3159--3185.

\bibitem[{Akujuobi and Zhang(2017)}]{akujuobi2017delve}
Uchenna Akujuobi and Xiangliang Zhang. 2017.
\newblock Delve: a dataset-driven scholarly search and analysis system.
\newblock \emph{ACM SIGKDD Explorations Newsletter}, 19(2):36--46.

\bibitem[{Athar(2011)}]{athar2011sentiment}
Awais Athar. 2011.
\newblock Sentiment analysis of citations using sentence structure-based features.
\newblock In \emph{Proceedings of the ACL 2011 student session}, pages 81--87.

\bibitem[{Athar and Teufel(2012)}]{athar-teufel-2012-context}
Awais Athar and Simone Teufel. 2012.
\newblock \href {https://aclanthology.org/N12-1073} {Context-enhanced citation sentiment detection}.
\newblock In \emph{Proceedings of the 2012 Conference of the North {A}merican Chapter of the Association for Computational Linguistics: Human Language Technologies}, pages 597--601, Montr{\'e}al, Canada. Association for Computational Linguistics.

\bibitem[{Beltagy et~al.(2020)Beltagy, Peters, and Cohan}]{Beltagy2020Longformer}
Iz~Beltagy, Matthew~E. Peters, and Arman Cohan. 2020.
\newblock Longformer: The long-document transformer.
\newblock \emph{arXiv:2004.05150}.

\bibitem[{Cao et~al.(2018)Cao, Wei, Li, and Li}]{cao2018faithful}
Ziqiang Cao, Furu Wei, Wenjie Li, and Sujian Li. 2018.
\newblock Faithful to the original: Fact aware neural abstractive summarization.
\newblock In \emph{Proceedings of the AAAI Conference on Artificial Intelligence}, volume~32.

\bibitem[{Chen and Zhuge(2019)}]{chen2019automatic}
Jingqiang Chen and Hai Zhuge. 2019.
\newblock Automatic generation of related work through summarizing citations.
\newblock \emph{Concurrency and Computation: Practice and Experience}, 31(3):e4261.

\bibitem[{Chen et~al.(2022)Chen, Alamro, Li, Gao, Yan, Gao, and Zhang}]{chen2022target}
Xiuying Chen, Hind Alamro, Mingzhe Li, Shen Gao, Rui Yan, Xin Gao, and Xiangliang Zhang. 2022.
\newblock Target-aware abstractive related work generation with contrastive learning.
\newblock In \emph{Proceedings of the 45th International ACM SIGIR Conference on Research and Development in Information Retrieval}, pages 373--383.

\bibitem[{Chen et~al.(2021)Chen, Alamro, Li, Gao, Zhang, Zhao, and Yan}]{chen-etal-2021-capturing}
Xiuying Chen, Hind Alamro, Mingzhe Li, Shen Gao, Xiangliang Zhang, Dongyan Zhao, and Rui Yan. 2021.
\newblock \href {https://doi.org/10.18653/v1/2021.acl-long.473} {Capturing relations between scientific papers: An abstractive model for related work section generation}.
\newblock In \emph{Proceedings of the 59th Annual Meeting of the Association for Computational Linguistics and the 11th International Joint Conference on Natural Language Processing (Volume 1: Long Papers)}, pages 6068--6077, Online. Association for Computational Linguistics.

\bibitem[{Cohan et~al.(2019)Cohan, Ammar, Van~Zuylen, and Cady}]{cohan2019structural}
Arman Cohan, Waleed Ammar, Madeleine Van~Zuylen, and Field Cady. 2019.
\newblock Structural scaffolds for citation intent classification in scientific publications.
\newblock \emph{arXiv preprint arXiv:1904.01608}.

\bibitem[{Deng et~al.(2021)Deng, Zeng, Gu, Ji, and Hua}]{deng2021automatic}
Zekun Deng, Zixin Zeng, Weiye Gu, Jiawen Ji, and Bolin Hua. 2021.
\newblock Automatic related work section generation by sentence extraction and reordering.

\bibitem[{Dong and Sch{\"a}fer(2011)}]{dong2011ensemble}
Cailing Dong and Ulrich Sch{\"a}fer. 2011.
\newblock Ensemble-style self-training on citation classification.
\newblock In \emph{Proceedings of 5th international joint conference on natural language processing}, pages 623--631.

\bibitem[{Erkan and Radev(2004)}]{erkan2004lexrank}
G{\"u}nes Erkan and Dragomir~R Radev. 2004.
\newblock Lexrank: Graph-based lexical centrality as salience in text summarization.
\newblock \emph{Journal of artificial intelligence research}, 22:457--479.

\bibitem[{Falke et~al.(2019)Falke, Ribeiro, Utama, Dagan, and Gurevych}]{falke-etal-2019-ranking}
Tobias Falke, Leonardo F.~R. Ribeiro, Prasetya~Ajie Utama, Ido Dagan, and Iryna Gurevych. 2019.
\newblock \href {https://doi.org/10.18653/v1/P19-1213} {Ranking generated summaries by correctness: An interesting but challenging application for natural language inference}.
\newblock In \emph{Proceedings of the 57th Annual Meeting of the Association for Computational Linguistics}, pages 2214--2220, Florence, Italy. Association for Computational Linguistics.

\bibitem[{Garfield et~al.(1965)}]{garfield1965can}
Eugene Garfield et~al. 1965.
\newblock Can citation indexing be automated.
\newblock In \emph{Statistical association methods for mechanized documentation, symposium proceedings}, volume 269, pages 189--192. Washington.

\bibitem[{Ge et~al.(2021)Ge, Dinh, Liu, Su, Lu, Wang, and Diesner}]{ge-etal-2021-baco}
Yubin Ge, Ly~Dinh, Xiaofeng Liu, Jinsong Su, Ziyao Lu, Ante Wang, and Jana Diesner. 2021.
\newblock \href {https://doi.org/10.18653/v1/2021.acl-long.116} {{BACO}: A background knowledge- and content-based framework for citing sentence generation}.
\newblock In \emph{Proceedings of the 59th Annual Meeting of the Association for Computational Linguistics and the 11th International Joint Conference on Natural Language Processing (Volume 1: Long Papers)}, pages 1466--1478, Online. Association for Computational Linguistics.

\bibitem[{Goodrich et~al.(2019)Goodrich, Rao, Liu, and Saleh}]{10.1145/3292500.3330955}
Ben Goodrich, Vinay Rao, Peter~J. Liu, and Mohammad Saleh. 2019.
\newblock \href {https://doi.org/10.1145/3292500.3330955} {Assessing the factual accuracy of generated text}.
\newblock In \emph{Proceedings of the 25th ACM SIGKDD International Conference on Knowledge Discovery \& Data Mining}, KDD '19, page 166–175, New York, NY, USA. Association for Computing Machinery.

\bibitem[{Grusky et~al.(2018)Grusky, Naaman, and Artzi}]{grusky-etal-2018-newsroom}
Max Grusky, Mor Naaman, and Yoav Artzi. 2018.
\newblock \href {https://doi.org/10.18653/v1/N18-1065} {{N}ewsroom: A dataset of 1.3 million summaries with diverse extractive strategies}.
\newblock In \emph{Proceedings of the 2018 Conference of the North {A}merican Chapter of the Association for Computational Linguistics: Human Language Technologies, Volume 1 (Long Papers)}, pages 708--719, New Orleans, Louisiana. Association for Computational Linguistics.

\bibitem[{Gu and Hahnloser(2023)}]{gu2023controllable}
Nianlong Gu and Richard H.~R. Hahnloser. 2023.
\newblock \href {http://arxiv.org/abs/2211.07066} {Controllable citation sentence generation with language models}.

\bibitem[{Hoang and Kan(2010)}]{hoang-kan-2010-towards}
Cong Duy~Vu Hoang and Min-Yen Kan. 2010.
\newblock \href {https://aclanthology.org/C10-2049} {Towards automated related work summarization}.
\newblock In \emph{Coling 2010: Posters}, pages 427--435, Beijing, China. Coling 2010 Organizing Committee.

\bibitem[{Hu and Wan(2014)}]{hu2014automatic}
Yue Hu and Xiaojun Wan. 2014.
\newblock Automatic generation of related work sections in scientific papers: an optimization approach.
\newblock In \emph{Proceedings of the 2014 Conference on Empirical Methods in Natural Language Processing (EMNLP)}, pages 1624--1633.

\bibitem[{Jaidka et~al.(2010)Jaidka, Khoo, and Na}]{jaidka2010imitating}
Kokil Jaidka, Christopher Khoo, and Jin-Cheon Na. 2010.
\newblock Imitating human literature review writing: An approach to multi-document summarization.
\newblock In \emph{International Conference on Asian Digital Libraries}, pages 116--119. Springer.

\bibitem[{Jaidka et~al.(2013{\natexlab{a}})Jaidka, Khoo, and Na}]{jaidka2013deconstructing}
Kokil Jaidka, Christopher Khoo, and Jin-Cheon Na. 2013{\natexlab{a}}.
\newblock Deconstructing human literature reviews--a framework for multi-document summarization.
\newblock In \emph{proceedings of the 14th European workshop on natural language generation}, pages 125--135.

\bibitem[{Jaidka et~al.(2013{\natexlab{b}})Jaidka, Khoo, and Na}]{jaidka2013literature}
Kokil Jaidka, Christopher~SG Khoo, and Jin-Cheon Na. 2013{\natexlab{b}}.
\newblock Literature review writing: how information is selected and transformed.
\newblock In \emph{Aslib Proceedings}. Emerald Group Publishing Limited.

\bibitem[{Jaidka et~al.(2011)Jaidka, Khoo, and Na}]{jaidka2011literature}
Kokil~Jaidka Jaidka, Christopher~Khoo Khoo, and Jin-Cheon~Na Na. 2011.
\newblock Literature review writing: a study of information selection from cited papers/kokil jaidka, christopher khoo and jin-cheon na.

\bibitem[{Jung et~al.(2022)Jung, Lin, Liao, Yuan, and Sun}]{jung2022intent}
Shing-Yun Jung, Ting-Han Lin, Chia-Hung Liao, Shyan-Ming Yuan, and Chuen-Tsai Sun. 2022.
\newblock Intent-controllable citation text generation.
\newblock \emph{Mathematics}, 10(10):1763.

\bibitem[{Jurgens et~al.(2018)Jurgens, Kumar, Hoover, McFarland, and Jurafsky}]{jurgens2018measuring}
David Jurgens, Srijan Kumar, Raine Hoover, Dan McFarland, and Dan Jurafsky. 2018.
\newblock Measuring the evolution of a scientific field through citation frames.
\newblock \emph{Transactions of the Association for Computational Linguistics}, 6:391--406.

\bibitem[{Khoo et~al.(2011)Khoo, Na, and Jaidka}]{khoo2011analysis}
Christopher~SG Khoo, Jin-Cheon Na, and Kokil Jaidka. 2011.
\newblock Analysis of the macro-level discourse structure of literature reviews.
\newblock \emph{Online Information Review}.

\bibitem[{Klavans et~al.(2001)Klavans, Kan, and McKeown}]{klavans2001domain}
Judith~L Klavans, Min-yen Kan, and Kathleen McKeown. 2001.
\newblock Domain-specific informative and indicative summarization for information retrieval.
\newblock \emph{Proceedings of the Document Understanding Workshop}.

\bibitem[{Knopf(2006)}]{knopf2006doing}
Jeffrey~W Knopf. 2006.
\newblock Doing a literature review.
\newblock \emph{PS: Political Science \& Politics}, 39(1):127--132.

\bibitem[{Kryscinski et~al.(2019)Kryscinski, Keskar, McCann, Xiong, and Socher}]{kryscinski-etal-2019-neural}
Wojciech Kryscinski, Nitish~Shirish Keskar, Bryan McCann, Caiming Xiong, and Richard Socher. 2019.
\newblock \href {https://doi.org/10.18653/v1/D19-1051} {Neural text summarization: A critical evaluation}.
\newblock In \emph{Proceedings of the 2019 Conference on Empirical Methods in Natural Language Processing and the 9th International Joint Conference on Natural Language Processing (EMNLP-IJCNLP)}, pages 540--551, Hong Kong, China. Association for Computational Linguistics.

\bibitem[{Lauscher et~al.(2022)Lauscher, Ko, Kuehl, Johnson, Cohan, Jurgens, and Lo}]{lauscher-etal-2022-multicite}
Anne Lauscher, Brandon Ko, Bailey Kuehl, Sophie Johnson, Arman Cohan, David Jurgens, and Kyle Lo. 2022.
\newblock \href {https://doi.org/10.18653/v1/2022.naacl-main.137} {{M}ulti{C}ite: Modeling realistic citations requires moving beyond the single-sentence single-label setting}.
\newblock In \emph{Proceedings of the 2022 Conference of the North American Chapter of the Association for Computational Linguistics: Human Language Technologies}, pages 1875--1889, Seattle, United States. Association for Computational Linguistics.

\bibitem[{Lauscher et~al.(2021)Lauscher, Ko, Kuhl, Johnson, Jurgens, Cohan, and Lo}]{lauscher2021multicite}
Anne Lauscher, Brandon Ko, Bailey Kuhl, Sophie Johnson, David Jurgens, Arman Cohan, and Kyle Lo. 2021.
\newblock Multicite: Modeling realistic citations requires moving beyond the single-sentence single-label setting.
\newblock \emph{arXiv preprint arXiv:2107.00414}.

\bibitem[{Lewis et~al.(2020)Lewis, Perez, Piktus, Petroni, Karpukhin, Goyal, K{\"u}ttler, Lewis, Yih, Rockt{\"a}schel et~al.}]{lewis2020retrieval}
Patrick Lewis, Ethan Perez, Aleksandra Piktus, Fabio Petroni, Vladimir Karpukhin, Naman Goyal, Heinrich K{\"u}ttler, Mike Lewis, Wen-tau Yih, Tim Rockt{\"a}schel, et~al. 2020.
\newblock Retrieval-augmented generation for knowledge-intensive nlp tasks.
\newblock \emph{Advances in Neural Information Processing Systems}, 33:9459--9474.

\bibitem[{Li et~al.(2023)Li, Lee, and Ouyang}]{li2023cited}
Xiangci Li, Yi-Hui Lee, and Jessica Ouyang. 2023.
\newblock Cited text spans for citation text generation.
\newblock \emph{arXiv preprint arXiv:2309.06365}.

\bibitem[{Li et~al.(2022)Li, Mandal, and Ouyang}]{li-etal-2022-corwa}
Xiangci Li, Biswadip Mandal, and Jessica Ouyang. 2022.
\newblock \href {https://doi.org/10.18653/v1/2022.naacl-main.397} {{CORWA}: A citation-oriented related work annotation dataset}.
\newblock In \emph{Proceedings of the 2022 Conference of the North American Chapter of the Association for Computational Linguistics: Human Language Technologies}, pages 5426--5440, Seattle, United States. Association for Computational Linguistics.

\bibitem[{Li and Ouyang(2024)}]{li2024explaining}
Xiangci Li and Jessica Ouyang. 2024.
\newblock Explaining relationships among research papers.
\newblock \emph{arXiv preprint arXiv:2402.13426}.

\bibitem[{Lin(2004)}]{lin2004rouge}
Chin-Yew Lin. 2004.
\newblock Rouge: A package for automatic evaluation of summaries.
\newblock In \emph{Text summarization branches out}, pages 74--81.

\bibitem[{Liu et~al.(2023)Liu, Zhang, Shi, Naseem, Wang, Hu, and Tsang}]{liu-etal-2023-causal}
Jiachang Liu, Qi~Zhang, Chongyang Shi, Usman Naseem, Shoujin Wang, Liang Hu, and Ivor Tsang. 2023.
\newblock \href {https://doi.org/10.18653/v1/2023.findings-emnlp.141} {Causal intervention for abstractive related work generation}.
\newblock In \emph{Findings of the Association for Computational Linguistics: EMNLP 2023}, pages 2148--2159, Singapore. Association for Computational Linguistics.

\bibitem[{Liu and Lapata(2019)}]{liu-lapata-2019-text}
Yang Liu and Mirella Lapata. 2019.
\newblock \href {https://doi.org/10.18653/v1/D19-1387} {Text summarization with pretrained encoders}.
\newblock In \emph{Proceedings of the 2019 Conference on Empirical Methods in Natural Language Processing and the 9th International Joint Conference on Natural Language Processing (EMNLP-IJCNLP)}, pages 3730--3740, Hong Kong, China. Association for Computational Linguistics.

\bibitem[{Lo et~al.(2020)Lo, Wang, Neumann, Kinney, and Weld}]{lo-wang-2020-s2orc}
Kyle Lo, Lucy~Lu Wang, Mark Neumann, Rodney Kinney, and Daniel Weld. 2020.
\newblock \href {https://doi.org/10.18653/v1/2020.acl-main.447} {{S}2{ORC}: The semantic scholar open research corpus}.
\newblock In \emph{Proceedings of the 58th Annual Meeting of the Association for Computational Linguistics}, pages 4969--4983, Online. Association for Computational Linguistics.

\bibitem[{Luu et~al.(2021)Luu, Wu, Koncel-Kedziorski, Lo, Cachola, and Smith}]{luu-etal-2021-explaining}
Kelvin Luu, Xinyi Wu, Rik Koncel-Kedziorski, Kyle Lo, Isabel Cachola, and Noah~A. Smith. 2021.
\newblock \href {https://doi.org/10.18653/v1/2021.acl-long.166} {Explaining relationships between scientific documents}.
\newblock In \emph{Proceedings of the 59th Annual Meeting of the Association for Computational Linguistics and the 11th International Joint Conference on Natural Language Processing (Volume 1: Long Papers)}, pages 2130--2144, Online. Association for Computational Linguistics.

\bibitem[{Mandal et~al.(2024)Mandal, Li, and Ouyang}]{mandal2024contextualizing}
Biswadip Mandal, Xiangci Li, and Jessica Ouyang. 2024.
\newblock Contextualizing generated citation texts.
\newblock \emph{arXiv preprint arXiv:2402.18054}.

\bibitem[{Martin-Boyle et~al.(2024)Martin-Boyle, Tyagi, Hearst, and Kang}]{martin2024shallow}
Anna Martin-Boyle, Aahan Tyagi, Marti~A Hearst, and Dongyeop Kang. 2024.
\newblock Shallow synthesis of knowledge in gpt-generated texts: A case study in automatic related work composition.
\newblock \emph{arXiv preprint arXiv:2402.12255}.

\bibitem[{Meng et~al.(2021)Meng, Thaker, Zhang, Dong, Yuan, Wang, and He}]{meng-etal-2021-bringing}
Rui Meng, Khushboo Thaker, Lei Zhang, Yue Dong, Xingdi Yuan, Tong Wang, and Daqing He. 2021.
\newblock \href {https://doi.org/10.18653/v1/2021.acl-short.137} {Bringing structure into summaries: a faceted summarization dataset for long scientific documents}.
\newblock In \emph{Proceedings of the 59th Annual Meeting of the Association for Computational Linguistics and the 11th International Joint Conference on Natural Language Processing (Volume 2: Short Papers)}, pages 1080--1089, Online. Association for Computational Linguistics.

\bibitem[{Mihalcea and Tarau(2004)}]{mihalcea2004textrank}
Rada Mihalcea and Paul Tarau. 2004.
\newblock Textrank: Bringing order into text.
\newblock In \emph{Proceedings of the 2004 conference on empirical methods in natural language processing}, pages 404--411.

\bibitem[{Narayan et~al.(2018)Narayan, Cohen, and Lapata}]{xsum-emnlp}
Shashi Narayan, Shay~B. Cohen, and Mirella Lapata. 2018.
\newblock Don't give me the details, just the summary! {T}opic-aware convolutional neural networks for extreme summarization.
\newblock In \emph{Proceedings of the 2018 Conference on Empirical Methods in Natural Language Processing}, Brussels, Belgium.

\bibitem[{Papineni et~al.(2002)Papineni, Roukos, Ward, and Zhu}]{papineni2002bleu}
Kishore Papineni, Salim Roukos, Todd Ward, and Wei-Jing Zhu. 2002.
\newblock Bleu: a method for automatic evaluation of machine translation.
\newblock In \emph{Proceedings of the 40th annual meeting of the Association for Computational Linguistics}, pages 311--318.

\bibitem[{Radev et~al.(2004)Radev, Jing, Sty{\'s}, and Tam}]{radev2004centroid}
Dragomir~R Radev, Hongyan Jing, Ma{\l}gorzata Sty{\'s}, and Daniel Tam. 2004.
\newblock Centroid-based summarization of multiple documents.
\newblock \emph{Information Processing \& Management}, 40(6):919--938.

\bibitem[{Radev et~al.(2013)Radev, Muthukrishnan, Qazvinian, and Abu-Jbara}]{radev2013acl}
Dragomir~R Radev, Pradeep Muthukrishnan, Vahed Qazvinian, and Amjad Abu-Jbara. 2013.
\newblock The acl anthology network corpus.
\newblock \emph{Language Resources and Evaluation}, 47(4):919--944.

\bibitem[{Ravi et~al.(2018)Ravi, Setlur, Ravi, and Govindaraju}]{ravi2018article}
Kumar Ravi, Srirangaraj Setlur, Vadlamani Ravi, and Venu Govindaraju. 2018.
\newblock Article citation sentiment analysis using deep learning.
\newblock In \emph{2018 IEEE 17th International Conference on Cognitive Informatics \& Cognitive Computing (ICCI* CC)}, pages 78--85. IEEE.

\bibitem[{See et~al.(2017)See, Liu, and Manning}]{see-etal-2017-get}
Abigail See, Peter~J. Liu, and Christopher~D. Manning. 2017.
\newblock \href {https://doi.org/10.18653/v1/P17-1099} {Get to the point: Summarization with pointer-generator networks}.
\newblock In \emph{Proceedings of the 55th Annual Meeting of the Association for Computational Linguistics (Volume 1: Long Papers)}, pages 1073--1083, Vancouver, Canada. Association for Computational Linguistics.

\bibitem[{Shuster et~al.(2021)Shuster, Poff, Chen, Kiela, and Weston}]{shuster-etal-2021-retrieval-augmentation}
Kurt Shuster, Spencer Poff, Moya Chen, Douwe Kiela, and Jason Weston. 2021.
\newblock \href {https://doi.org/10.18653/v1/2021.findings-emnlp.320} {Retrieval augmentation reduces hallucination in conversation}.
\newblock In \emph{Findings of the Association for Computational Linguistics: EMNLP 2021}, pages 3784--3803, Punta Cana, Dominican Republic. Association for Computational Linguistics.

\bibitem[{Teufel et~al.(2006)Teufel, Siddharthan, and Tidhar}]{teufel2006automatic}
Simone Teufel, Advaith Siddharthan, and Dan Tidhar. 2006.
\newblock Automatic classification of citation function.
\newblock In \emph{Proceedings of the 2006 conference on empirical methods in natural language processing}, pages 103--110.

\bibitem[{Tuarob et~al.(2019)Tuarob, Kang, Wettayakorn, Pornprasit, Sachati, Hassan, and Haddawy}]{tuarob2019automatic}
Suppawong Tuarob, Sung~Woo Kang, Poom Wettayakorn, Chanatip Pornprasit, Tanakitti Sachati, Saeed-Ul Hassan, and Peter Haddawy. 2019.
\newblock Automatic classification of algorithm citation functions in scientific literature.
\newblock \emph{IEEE Transactions on Knowledge and Data Engineering}, 32(10):1881--1896.

\bibitem[{Vaswani et~al.(2017)Vaswani, Shazeer, Parmar, Uszkoreit, Jones, Gomez, Kaiser, and Polosukhin}]{vaswani2017attention}
Ashish Vaswani, Noam Shazeer, Niki Parmar, Jakob Uszkoreit, Llion Jones, Aidan~N Gomez, {\L}ukasz Kaiser, and Illia Polosukhin. 2017.
\newblock Attention is all you need.
\newblock In \emph{Advances in neural information processing systems}, pages 5998--6008.

\bibitem[{Velickovic et~al.(2018)Velickovic, Cucurull, Casanova, Romero, Lio’, and Bengio}]{Velickovic2018GraphAN}
Petar Velickovic, Guillem Cucurull, Arantxa Casanova, Adriana Romero, Pietro Lio’, and Yoshua Bengio. 2018.
\newblock Graph attention networks.
\newblock \emph{ArXiv}, abs/1710.10903.

\bibitem[{Wang et~al.(2019)Wang, Li, Zhou, Tang, and Wang}]{wang2019toc}
Pancheng Wang, Shasha Li, Haifang Zhou, Jintao Tang, and Ting Wang. 2019.
\newblock Toc-rwg: Explore the combination of topic model and citation information for automatic related work generation.
\newblock \emph{IEEE Access}, 8:13043--13055.

\bibitem[{Wang et~al.(2018)Wang, Liu, and Gao}]{wang-etal-2018-neural-related}
Yongzhen Wang, Xiaozhong Liu, and Zheng Gao. 2018.
\newblock \href {https://doi.org/10.18653/v1/D18-1204} {Neural related work summarization with a joint context-driven attention mechanism}.
\newblock In \emph{Proceedings of the 2018 Conference on Empirical Methods in Natural Language Processing}, pages 1776--1786, Brussels, Belgium. Association for Computational Linguistics.

\bibitem[{Wasson(1998)}]{wasson1998using}
Mark Wasson. 1998.
\newblock Using leading text for news summaries: Evaluation results and implications for commercial summarization applications.
\newblock In \emph{36th Annual Meeting of the Association for Computational Linguistics and 17th International Conference on Computational Linguistics, Volume 2}, pages 1364--1368.

\bibitem[{Xing et~al.(2020)Xing, Fan, and Wan}]{xing2020automatic}
Xinyu Xing, Xiaosheng Fan, and Xiaojun Wan. 2020.
\newblock Automatic generation of citation texts in scholarly papers: A pilot study.
\newblock In \emph{Proceedings of the 58th Annual Meeting of the Association for Computational Linguistics}, pages 6181--6190.

\bibitem[{Yasunaga et~al.(2019)Yasunaga, Kasai, Zhang, Fabbri, Li, Friedman, and Radev}]{yasunaga2019scisummnet}
Michihiro Yasunaga, Jungo Kasai, Rui Zhang, Alexander~R Fabbri, Irene Li, Dan Friedman, and Dragomir~R Radev. 2019.
\newblock Scisummnet: A large annotated corpus and content-impact models for scientific paper summarization with citation networks.
\newblock In \emph{Proceedings of the AAAI Conference on Artificial Intelligence}, volume~33, pages 7386--7393.

\bibitem[{Zhao et~al.(2019)Zhao, Luo, Feng, Zheng, and Liu}]{zhao2019context}
He~Zhao, Zhunchen Luo, Chong Feng, Anqing Zheng, and Xiaopeng Liu. 2019.
\newblock A context-based framework for modeling the role and function of on-line resource citations in scientific literature.
\newblock In \emph{Proceedings of the 2019 Conference on Empirical Methods in Natural Language Processing and the 9th International Joint Conference on Natural Language Processing (EMNLP-IJCNLP)}, pages 5206--5215.

\end{thebibliography}
\clearpage
\appendix

\section{Appendix}
\label{sec:appendix}

\begin{table*}[t]
\begin{center}
    \begin{tabular}{ | p{0.25\linewidth}|p{0.55\linewidth} | }
    \hline
    \textbf{Prior Work} & \textbf{Inputs} \\ \hline
    \citet{hoang-kan-2010-towards} & Topic hierarchy tree of the target related work, full cited papers \\  \hline 
    \citet{hu2014automatic} & Target paper (abstract, introduction), cited papers (abstract, introduction, related work, conclusion)\\ \hline
    \citet{wang-etal-2018-neural-related} & Full-texts of cited papers\\ \hline
    \citet{chen2019automatic} &  Title, abstract, introduction, and conclusion for both target paper and cited papers; papers that co-cite the cited papers \\ \hline
    \citet{wang2019toc} & Full papers of target paper and cited papers, citation sentences that co-citing the cited papers\\ \hline
    \citet{deng2021automatic} & Abstract or conclusion sections of the cited papers \\ \hline
    \end{tabular}
    \caption{A summary of the problem formulations of the prior works on extractive related work generation. All of their generation targets are a sequence of extracted sentences.} \label{tab:task_formulation_extractive}
    \vspace{-1em}
\end{center}
\end{table*}

\begin{table*}[t]
\begin{center}
    \begin{tabular}{ | p{0.25\linewidth}|p{0.35\linewidth} | p{0.35\linewidth} | }
    \hline
    \textbf{Prior Work} & \textbf{Inputs} & \textbf{Target} \\ \hline
    \citet{abura2020automatic} & Cited title, abstract & Citation sentence w/ single reference \\ \hline
    \citet{xing2020automatic} & Context sentences, single cited abstract & Citation sentence w/ single  reference\\ \hline
    \citet{ge-etal-2021-baco} & Citation network, single cited abstract, context sentences & Citation sentence, citation function, salient sentence in cited abstracts\\ \hline
    \citet{luu-etal-2021-explaining} & Intro of the citing paper, named entities of the cited papers & Citation sentence w/ single reference \\ \hline
    \citet{li-etal-2022-corwa} & Context sentences w/o the target span, 1+ cited abstracts & Citation span w/ 1+ citations \\ \hline
    \citet{jung2022intent} & Abstract or title of the citing paper, cited abstract, citation intent & 1+ citation sentences with single citation \\ \hline
    \citet{li2023cited} & Context sentences w/o the target span, 1+ cited abstracts & Citation span w/ 1+ citations \\ \hline 
    \citet{gu2023controllable} & Title, abstract of the target paper \& cited paper; citation text, citation intent, keywords & Citation sentence with presumably single citation\\ \hline
    \citet{mandal2024contextualizing}  & Context sentences w/o the target span, 1+ cited abstracts & Context sentences w/ the target span w/ 1+ citations \\ \hline
    \hline
    \citet{chen-etal-2021-capturing} & Cited abstracts & A paragraph w/ 2+ citations \\ \hline
    \citet{chen2022target} & Target abstract, cited abstracts & A paragraph w/ 2+ citations \\ \hline
    \citet{liu-etal-2023-causal} & Cited abstracts & Related work paragraph \\ \hline
    \citet{li2024explaining} & Main idea of the RWS, title, abstract, intro, conclusion of the target paper, full text of cited papers & 1+ paragraphs of RWS \\ \hline
    \citet{martin2024shallow} & Target paper w/o RWS, and abstracts of the cited papers & RWS \\ \hline
    \end{tabular}
    \caption{A summary of the problem formulations of the prior works on neural network-based related work generation. ``Context'' refers to those sentences or paragraphs around the target citation sentences.} \label{tab:task_formulation_abstractive}
    \vspace{-1em}
\end{center}
\end{table*}

\begin{table*}[t]
\begin{minipage}{\textwidth}
\begin{center}
    \begin{tabular}{ | p{0.25\linewidth}|p{0.3\linewidth} | p{0.25\linewidth} | p{0.12\linewidth} |}
    \hline
    \textbf{Prior Work} & \textbf{Source Domain} & \textbf{Size} & \textbf{Dataset} \\ \hline
    \citet{hoang-kan-2010-towards} & Papers from NLP and IR, manually curated topic tree & 20 papers & RWSData \footnote{\url{http://wing.comp.nus.edu.sg/downloads/rwsdata}} \\ \hline 
    \citet{hu2014automatic} & ACL Anthology & 1050 papers & N/A \\ \hline
    \citet{wang-etal-2018-neural-related} & ACM digital library & 8080 papers & Available \footnote{\url{https://github.com/kuadmu/2018EMNLP}} \\ \hline
    \citet{chen2019automatic} & ACL Anthology \& IJCAI & 25 papers & RWS-Cit \footnote{\url{https://github.com/jingqiangchen/RWS-Cit}} \\\hline
    \citet{wang2019toc} & NLP conferences & 50 papers & NudtRwG \footnote{\url{https://github.com/ NudtRwG/NudtRwG-Dataset}} \\ \hline
    \citet{deng2021automatic} & ScisummNet (ACL) & 11954 examples & N/A \\ \hline
    \hline
    \citet{abura2020automatic} & ScisummNet (ACL) & 940 + 15574 pairs & N/A \\
    \hline
    \citet{xing2020automatic} & ACL Anthology Network & 1k + 85k examples & Available \footnote{\url{https://github.com/XingXinyu96/citation_generation}} \\ \hline
    \citet{ge-etal-2021-baco} & ACL Anthology Network & 1.2k + 84k examples & N/A \\ \hline \hline

    \citet{luu-etal-2021-explaining} & S2ORC (CS) & 622k citations from 154k papers & Extraction from S2ORC \footnote{\url{https://github.com/Kel-Lu/SciGen/tree/master/data_processing}}\\ \hline
    \citet{li-etal-2022-corwa} & S2ORC (NLP) & Annotated 3565 dominant spans \& 4228 reference spans from 2927 paragraphs; 565+362+11465 train/test/distant RWS & CORWA \footnote{\url{https://github.com/jacklxc/CORWA}} \\ \hline
    \citet{jung2022intent} & SciCite (CS) & 8243/916/1861 train/validation/test samples & Available \footnote{\url{https://github.com/BradLin0819/Automatic-Citation-Text-Generation-with-Citation-Intent-Control}} \\ \hline
    \citet{li2023cited} & CORWA, S2ORC (NLP) & 1654/1206/19784 train/test/distant dominant citation spans & Available \footnote{\url{https://github.com/jacklxc/CTS4CitationTextGeneration}} \\ \hline 
    \citet{gu2023controllable} & arXiv computer science papers & 233.6k/1.3k/1.1k train/validation/test samples & Available \footnote{\url{https://github.com/nianlonggu/LMCiteGen}}\\ \hline 
    \citet{mandal2024contextualizing} & CORWA &  565/362/11465 train/test/distant RWS & N/A \\ \hline
    \hline
    \citet{chen-etal-2021-capturing} & S2ORC (Multi-domain), Delve (CS) & 150k, 80k examples & Available \footnote{\url{https://github.com/iriscxy/relatedworkgeneration}} \\ \hline
    \citet{chen2022target} & S2ORC (Multi-domain), Delve (CS) & 107.7k/5k/5k, 208.3k/5k/5k train/dev/test examples & N/A \\ \hline
    \citet{liu-etal-2023-causal} & S2ORC (Multi-domain), Delve (CS) & 126k/5k/5k, 72k/3k/3k train/dev/test pairs & N/A \\ \hline
    \citet{li2024explaining} & PDFs from NLP, ML, Speech, CV, etc. & 38 papers & N/A \\ \hline
    \citet{martin2024shallow} & 2023 ACL best papers & 10 papers & N/A \\ \hline
    \end{tabular}
    \caption{A summary of the datasets of the prior works on related work generation.} \label{tab:datasets}
    \vspace{-1em}
\end{center}
\end{minipage}
\end{table*}

\begin{table*}[t]
\begin{center}
    \begin{tabular}{ | p{0.25\linewidth}|p{0.75\linewidth} | }
    \hline
    \textbf{Prior Work} & \textbf{Approaches} \\ \hline
    \citet{hoang-kan-2010-towards} & Heuristic approach to generate general and specific content separately given a topic tree \\ \hline 
    \citet{hu2014automatic} & PLSA for topic modeling, SVR for sentence importance score, and global optimization for sentence selection \\ \hline
    \citet{wang-etal-2018-neural-related} & Custom neural seq2seq model (CNN, LSTM, attention), random walk for encoding heterogeneous bibliography graph \\ \hline
    \citet{chen2019automatic} & Considering papers co-cite the cited papers; Representing graph for relationship modeling of papers, then finding sentence nodes that cover the minimum Steiner tree of the graph \\ \hline
    \citet{wang2019toc} & Leveraging both topic model and cited text spans \\ \hline 
    \citet{deng2021automatic} & BERT-based sentence extraction \& reordering \\ \hline
    \hline
    \citet{abura2020automatic} & Applying PTGen and Transformer \\ \hline
    \citet{xing2020automatic} & Manual annotation + automatic annotation of citation sentences; PTGEN-Cross based on cross-attention mechanism \\ \hline
    \citet{ge-etal-2021-baco} & Citation network as auxiliary input; citation function \& salient sentences in cited papers as auxiliary output; multi-task learning \\ \hline
    \citet{luu-etal-2021-explaining} & SciGPT2; IE-Extracted Term Lists; ranking based on entity matching \\ \hline 
    \citet{li-etal-2022-corwa} & LED-based citation span generation \\ \hline
    \citet{jung2022intent} & BART/T5-based citation sentence generation with citation intents \\ \hline
    \citet{li2023cited} & RAG \& LED; ROUGE-based CTS retrieval \\ \hline
    \citet{gu2023controllable} & Fine-tuned GPT-Neo \& Galactica with Proximal Policy Optimization \\ \hline
    \citet{mandal2024contextualizing} & Using citation context along with citation spans as generation target \\ \hline
    \hline
    \citet{chen-etal-2021-capturing} & Transformer-based hierarchical encoder; relationship modeling module\\ \hline
    \citet{chen2022target} & Improved over \citet{chen-etal-2021-capturing} by encoding target paper's abstract \\ \hline
    \citet{liu-etal-2023-causal} & Proposed a custom Causal Intervention Module (CaM) inserted between Transformer blocks \\ \hline
    \citet{li2024explaining} & GPT-3.5 for feature generation, e.g. faceted summary, relationship \& usage of citations; GPT-4 based RWG \\ \hline
    \citet{martin2024shallow} & GPT-4 with human-in-the-loop \\ \hline
    \end{tabular}
    \caption{A summary of the approaches of the prior works.} \label{tab:approaches}
    \vspace{-1em}
\end{center}
\end{table*}

\begin{table*}[t]
\begin{center}
    \begin{tabular}{| p{0.25\linewidth}|p{0.25\linewidth} | p{0.2\linewidth} | p{0.25\linewidth} | }
    \hline
    \textbf{Prior Work} & \textbf{Baselines} & \textbf{Automatic} & \textbf{Human Evaluation} \\ \hline
    \citet{hoang-kan-2010-towards} & LEAD, MEAD & ROUGE recall (1, 2, S4, SU4) & Correctness, novelty, fluency, usefulness \\\hline 
    \citet{hu2014automatic} & MEAD, LexRank & ROUGE F1 (1, 2, SU4)& Correctness, readability, usefulness\\\hline 
    \citet{wang-etal-2018-neural-related} & Luhn, MMR, LexRank, SumBasic, NltkSum, Pointer Network& ROUGE F1 (1, 2, L) & Compliance to target paper, intuitiveness, usefulness \\ \hline
    \citet{chen2019automatic} & MEAD, LexRank, RoWoS & ROUGE F1 (1, 2) & N/A \\ \hline
    \citet{wang2019toc} & LexRank, SumBasic, JS-Gen, TopicSum & ROUGE recall \& F1 (1, 2, SU4) & N/A \\\hline 
    \citet{deng2021automatic} & MEAD & ROUGE precision, recall, F1 (1, 2, L) & informativeness, fluency, succinctness\\ \hline

    \end{tabular}
    \caption{A summary of the evaluation methods of the extractive related work generation works.} \label{tab:extractive_evaluation}
    \vspace{-1em}
\end{center}
\end{table*}

\begin{table*}[t]
\begin{center}
    \begin{tabular}{| p{0.25\linewidth}|p{0.25\linewidth} | p{0.2\linewidth} | p{0.25\linewidth} | }
    \hline
    \textbf{Prior Work} & \textbf{Baselines} & \textbf{Automatic} & \textbf{Human Evaluation} \\ \hline
    \citet{abura2020automatic} & MEAD, TextRank, SUMMA, SEQ$^3$& ROUGE precision, recall, F1 (1, 2, L, SU4) & N/A \\ \hline
    \citet{xing2020automatic} & RandomSen, MaxSimSen, EXT-Oracle, COPY-CIT, PTGEN& ROUGE F1 (1, 2, L) & Readability, Content, Coherence, Overall \\ \hline
    \citet{ge-etal-2021-baco}& LexRank, TextRank, EXT-Oracle, PTGEN, PTGEN-Cross&ROUGE F1 (1, 2, L)& Fluency, relevance, coherence, overall\\ \hline
    \citet{luu-etal-2021-explaining} & N/A & BLEU, ROUGE-L & Correct, Specific \\ \hline 
    \citet{li-etal-2022-corwa} & Citation sentence generation & ROUGE F1 (1, 2, L) & Fluency, coherence, relevance, overall \\ \hline
    \citet{jung2022intent} & EXT-Oracle, ablations & ROUGE F1 (1, 2, L), SciBERTScore, citation intent accuracy & Correct, specific, plausible, intent \\ \hline
    \citet{li2023cited} & Citation span generation based on cited abstracts, \& human-annotated CTS & BLEU, ROUGE-F1-L, METEOR, QuestEval, ANLI & Fluency, coherence, relevance, overall \\ \hline
    \citet{gu2023controllable} & BART-base \& -large, GPT-Neo 125M \& 1.3B, Galactica 125M \& 1.3B \&6.7B, LLaMA-7B ablations, GPT-3.5-turbo & ROUGE F1 (1, 2, L), Intent alignment score, keyword recall, fluency score & Intent alignment, keyword recall, fluency \& similarity to the ground truth \\ \hline
    \citet{mandal2024contextualizing} & Ablations & N/A & Fluency, coherence, relevance, overall \\ \hline
    \hline
    \citet{chen-etal-2021-capturing} & LEAD, TextRank, BertSumEXT, MGSum-ext, PTGen+Cov, TransformerABS, BertSumAbs, MGSum-abs, GS & ROUGE F1 (1, 2, L)& QA, informativeness, coherence, succinctness \\ \hline
    \citet{chen2022target} & LEAD, LexRank, NES, BertSumEXT, MGSum, EMS, RRG, BertSumAbs & ROUGE F1 (1, 2, L, SU)& QA, informativeness, coherence, succinctness \\ \hline
    \citet{liu-etal-2023-causal} & TexRank, BertSumEXT, MGSum-ext \& -abs, TransformerABS, RRG, BertSumAbs, GS, T5-base, BART-base, Longformer, NG-abs, TAG & ROUGE F1 (1, 2, L)& QA, informativeness, coherence, succinctness \\ \hline
    \citet{li2024explaining} & Ablations & ROUGE F1 (1, 2, L) & Fluency, coherence, relevance (cited, target), factuality, usefulness, writing, overall, \# of errors \\ \hline 
    \citet{martin2024shallow} & Human \& Human-in-the-loop & \# of edges, average node degree, density, cluster coefficient & Qualitative analysis \\ \hline
    \end{tabular}
    \caption{A summary of the evaluation methods of the abstractive related work generation works.} \label{tab:abstractive_evaluation}
    \vspace{-1em}
\end{center}
\end{table*}

\begin{table*}[t]
\begin{center}
    \begin{tabular}{ | p{0.15\linewidth}|p{0.45\linewidth}|p{0.35\linewidth} | }
    \hline
    \textbf{Perspective} & \textbf{Definition} & \textbf{Used By} \\ \hline
    Fluency, Readability & Does the summary's exposition flow well, in terms of syntax as well as discourse? & \citet{hoang-kan-2010-towards, hu2014automatic, deng2021automatic, xing2020automatic, ge-etal-2021-baco, chen-etal-2021-capturing, chen2022target, li-etal-2022-corwa, li2023cited, gu2023controllable, mandal2024contextualizing, li2024explaining} \\\hline 
    Correctness & Is the summary content relevant to (express the factual relationship with) the hierarchical topics/cited papers given? & \citet{hoang-kan-2010-towards, hu2014automatic, luu-etal-2021-explaining, jung2022intent} \\ \hline
    Novelty & Does the summary introduce novel information that is significant in comparison with the human created summary? & \citet{hoang-kan-2010-towards}\\\hline
    Usefulness & Is the summary useful in supporting the researchers to quickly grasp the related works given hierarchical topics? & \citet{hoang-kan-2010-towards, hu2014automatic, wang-etal-2018-neural-related, li2024explaining} \\\hline
    Content, Relevance & Whether the citation text is relevant to the cited paper's abstract& \citet{wang-etal-2018-neural-related, xing2020automatic, ge-etal-2021-baco, li-etal-2022-corwa, li2023cited, gu2023controllable, mandal2024contextualizing, li2024explaining} \\\hline
    Coherence & Whether the citation text is coherent with the citing paper's context & \citet{xing2020automatic, ge-etal-2021-baco, chen-etal-2021-capturing, chen2022target, li-etal-2022-corwa, li2023cited, liu-etal-2023-causal, mandal2024contextualizing, li2024explaining} \\\hline
    Informativeness & Does the related work convey important facts about the topic question? & \citet{deng2021automatic, chen-etal-2021-capturing, chen2022target, liu-etal-2023-causal} \\ \hline
    Succinctness & Does the related work avoid repetition? & \citet{deng2021automatic, chen-etal-2021-capturing, liu-etal-2023-causal} \\ \hline
    Overall & Overall quality & \citet{xing2020automatic, ge-etal-2021-baco, li-etal-2022-corwa, li2023cited, mandal2024contextualizing, li2024explaining}\\\hline
    Intuitiveness & How intuitive is the related work section for readers to grasp the key content? & \citet{wang-etal-2018-neural-related} \\ \hline
    QA & Retain the key information? & \citet{chen-etal-2021-capturing, chen2022target, liu-etal-2023-causal} \\ \hline
    Specific & Whether the explanation describes a specific relationship between the two works & \citet{luu-etal-2021-explaining, jung2022intent} \\ \hline
    Factuality, \# of errors & Does the output contain factual errors? & \citet{li2024explaining} \\ \hline
    Plausible, writing & Writing style of citation text / RWS & \citet{jung2022intent, li2024explaining} \\ \hline
    Qualitative analysis & Descriptive case study & \citet{li2024explaining, martin2024shallow} \\ \hline
    Intent alignment & Whether the output aligns with the input intent. & \citet{jung2022intent, gu2023controllable} \\ \hline
    Keyword recall & Whether the output contains the input key words. & \citet{gu2023controllable} \\ \hline
    \end{tabular}
    \caption{A summary of the perspectives for human evaluation.} \label{tab:human_evaluation}
    \vspace{-1em}
\end{center}
\end{table*}

\end{document}